\documentclass[final]{l4dc2026}

\usepackage{amsmath}
\usepackage{amsfonts}
\usepackage{bm}
\usepackage{graphicx} 
\usepackage[skip=3.0pt]{caption}
\usepackage{subcaption}
\usepackage{wrapfig}
\usepackage{booktabs} % for professional tables
\usepackage{multirow}
\usepackage{makecell}
\usepackage{tablefootnote}
\usepackage{color}

\usepackage[super]{nth}
\usepackage{algorithm}
\usepackage{algpseudocode}
\usepackage{markdown}
\title[LaLQR]{Latent Linear Quadratic Regulator for Robotic Control Tasks}
\usepackage{times}

\author{%
    \Name{Yuan Zhang} \Email{yzhang@cs.uni-freiburg.de} \\
    \addr University of Freiburg, Germany
\AND
    \Name{Shaohui Yang} \Email{shaohui.yang@epfl.ch}\\
    \addr EPFL, Switzerland 
\AND
    \Name{Toshiyuki Ohtsuka} \Email{ohtsuka@i.kyoto-u.ac.jp}\\
    \addr Kyoto University, Japan 
\AND
    \Name{Colin Jones} \Email{colin.jones@epfl.ch}\\
    \addr EPFL, Switzerland 
\AND
  \Name{Joschka Boedecker} \Email{jboedeck@cs.uni-freiburg.de}\\
  \addr University of Freiburg, Germany
}

\begin{document}
\maketitle

%===============================================================================
\begin{abstract}
    Model predictive control (MPC) offers high-performance control but remains computationally expensive for nonlinear dynamics, hindering its real-time deployment in robotic tasks. Inspired by the Koopman operator, we propose the $\textbf{la}$tent $\textbf{l}$inear $\textbf{q}$uadratic $\textbf{r}$egulator (LaLQR) framework, which learns an alternative latent linear-quadratic structure enabling efficient LQR-based control for nonlinear systems. LaLQR enforces the fixed Brunovsky canonical form on the latent linear dynamics to ensure controllability and numerical stability, while jointly learning a nonlinear embedding and cost function under latent state and cost prediction objectives.
    Experiments on diverse MuJoCo simulated robotic tasks show that LaLQR achieves comparable control quality to expensive gradient-based optimization methods while offering superior computational efficiency and better generalization over learning-based baselines.
\end{abstract}

\begin{keywords}
    Model Predictive Control, Linear Quadratic Regulator, Imitation Learning, Robotics
\end{keywords}

%===============================================================================

% === 1.5 pages ===
\section{Introduction}

    % MPC and its complexity
    Model predictive control (MPC) has achieved remarkable success in robotic applications such as quadruped locomotion~\citep{dicarlo2018dynamic, grandia2023perceptive} and agile drone racing~\citep{song2020learning}. As a model-based control method, MPC optimizes future control sequences by predicting system evolution under a dynamic model $x_{h+1} = f_d(x_h, u_h)$, where $x$ and $u$ denote the system state and control input, respectively. However, the computational burden of solving MPC problems—particularly for nonlinear high-dimensional dynamics—remains a major obstacle to real-time deployment on embedded robotic platforms, e.g., humanoids~\citep{katayama2023model}.

    % Current method to reduce the complexity
    Various approaches have been proposed to reduce the complexity of nonlinear MPC. Sequential quadratic programming (SQP)~\citep{nocedal1999sequential} iteratively solves quadratic subproblems based on first- and second-order expansions of the nonlinear model and cost function, but the iterative nature of SQP often makes it too slow for real-time control. Local linearization methods, such as the localized LQR (LoLQR)~\citep{wang2014localized}, approximate the nonlinear system around an equilibrium point using a Taylor expansion and then apply the efficient LQR solver~\citep{anderson2007optimala}. However, their performance quickly degrades when the system operates far from the linearization point. More recently, imitation learning (IL)~\citep{brohan2022rt1} has been employed to bypass explicit optimization altogether by directly learning a mapping from states to actions using neural networks. Although IL can achieve excellent performance given sufficient expert data, it often exhibits poor generalization outside the training distribution.
    
    % Our method
    In this work, we propose the $\textbf{la}$tent $\textbf{l}$inear $\textbf{q}$uadratic $\textbf{r}$egulator (LaLQR), a method that combines the efficiency of LQR with the flexibility of nonlinear representation inspired by the Koopman operator framework. Specifically, LaLQR employs a nonlinear embedding $\phi$ that maps the original state $x_h$ into a latent state $z_h = \phi(x_h)$, where the dynamics evolve linearly and the cost is quadratic. 
    One of our key contributions is the adoption of the Brunovsky canonical form~\citep{brunovsky1970classification} to represent latent linear dynamics, owing to its extreme simplicity in structure and superior numerical stability during training.
    % We specifically apply the Brunovsky canonical form~\citep{brunovsky1970classification} for the linear dynamics without loss of generality, while maintaining numerical stability during training. 
    Therefore, only the embedding and cost functions are jointly learned by predicting the next latent state and cost. Subsequently, the LQR method is applied to compute the gain matrix, which, in conjunction with the embedding function, is utilized for efficient online control.
    Compared with learning-based methods such as IL, LaLQR preserves interpretability, stability, and generalization while reducing computation by an order of magnitude relative to SQP-based MPC. Experiments on multiple MuJoCo robotic control tasks demonstrate that LaLQR achieves a favorable trade-off among control performance, efficiency, and generalization.

% === 0.7 pages ===
\section{Related Work}
\label{sec:related_work}

% koopman operator
\noindent\textbf{The Koopman Operator for Control.}
    The Koopman operator has become increasingly popular in the control community due to its ability to transform the original state $x$ into the latent state $z$, in which the system dynamics are linear in $z$ and the control input $u$. This latent space can potentially assist the theoretical analysis and simplify the control tasks. To leverage the Koopman operator for optimal control, three components are typically required: (1) an embedding function $\phi$ that maps $x\!\mapsto\!z$ (some works~\citep{mondal2024efficient} also include a control mapping $u\!\mapsto\!v$); (2) a linear dynamic model parameterized by $(A,B)$; (3) a stage cost function. Both (2) and (3) operate in the latent space. These parameters can be learned in data-driven techniques such as extended dynamic mode decomposition (eDMD)~\citep{lusch2018deep}. The learning principle of eDMD can be summarized as $\textit{latent state prediction}$ (LSP), which aligns $\phi(x_{h+1})$ and $A\phi(x_h)+Bu_h$, where $h$ is the step index. Other learning principles have been proposed in the literature: (1) predicting the true cost from the latent state ($\textit{cost prediction}$, CP)~\citep{li2020causal}; (2) reconstructing the original state $x$ from the latent representation ($\textit{state reconstruction}$, SR)~\citep{watter2015embed}; (3) learning the cost-to-go function for each latent state($\phi_{V^*}$)~\citep{mondal2024efficient}. Notably, these learning principles all shape the structure of the latent space $z$ through gradient-based learning, effectively imposing different levels of abstraction on the original state $x$. Representative Koopman-based control methods are summarized in Table~\ref{tab:motivation}.

    % ------------------------
    % method comparison
    \begin{table*}[ht]
    \begin{center}
    	 \scalebox{0.7}{
    		\begin{tabular}{lccccc}
    			\toprule
    		  \textbf{Method} &\textbf{Embedding Function} &\textbf{Dynamic Model} & \textbf{Cost Function} & \textbf{Learning Principle} & \textbf{Optimization}  \\
    			\midrule	
    \citet{watter2015embed} & NN  & $A_tz+B_tu$ & $z^T\bar{Q}z+u^T\bar{R}u$  & LSP, SR, reg. & iLQG \\
    % \citet{abraham2017modelbased} & features & \red{$A_tx_t+B_tu$} & $ x_t^T\bar{Q}x_t$ & ZP & SAC (MPC) \\
    \citet{korda2018linear} & BF & $Az+Bu$ & $z^T\bar{Q}z+u^T\bar{R}u$  & LSP & QP \\
    \citet{bruder2019modeling} & BF & $Az+Bu$ & $z^T\bar{Q}_tz+u^T\bar{R}_tu$ & LSP, reg. & QP \\
    \citet{mamakoukas2019local} & BF & $Az+Bu$ & $z^T\bar{Q}z+u^T\bar{R}u$ & LSP & LQR \\
    \citet{li2020learningb} & NN & $Az+Bu$ & $z^T\bar{Q}z+u^T\bar{R}u$ & SR, reg. & QP \\
    \citet{yin2022embedding} & NN & $Az+Bu$  & $z^TQz+u^TRu$ & LSP, SR, $\phi_{V^*}$, reg. & LQR \\ 
    \citet{mondal2024efficient} & NN & $Az+Bv$  & NN  & LSP, CP, $\phi_{V^*}$ & MPPI \\
    \midrule
    % TDMPC~\citep{hansen2022temporal} & $f_T(z, u)$ & $f_C(z, u)$ &  ZP, CP, $\phi_{Q^*}$  & MPPI  \\
    % \midrule
     LaLQR (Ours) & NN & $\bar{A}_bz+\bar{B}_bv$\footnotemark & $z^TQz+v^TRv$  & LSP, CP & LQR \\
    			\bottomrule
    		\end{tabular}
    		}
    	\end{center}
    \vspace{-1.0em}
    \caption{Representative Koopman-based control methods. NN is short for neural network. BF is short for basis function. $\bar{X}$ means the module $X$ is fixed during training and $X_t$ means $X$ is time-variant. reg. means the additional regularizaiton in latent space.}
    \label{tab:motivation}
    \end{table*}
    \footnotetext{The Brunovsky canonical form as in Equation~\ref{eq:canonical-form-structure}.}
    % ------------------------
    
    Previous works~\citep{korda2018linear, bruder2019modeling, mamakoukas2019local} heavily rely on prior knowledge of the embedding function (basis function) and cost matrices $(Q,R)$, and primarily focus on learning the dynamic model (the Koopman operator) using the LSP principle. In contrast, our approach enables learnable embedding functions and cost matrices, thereby minimizing the required prior knowledge. A key feature of our method is the use of the Brunovsky canonical form $(\bar{A}_b, \bar{B}_b)$~\citep{brunovsky1970classification} for the latent dynamics. The fixed matrices maintain expressiveness through the nonlinear embedding while enhancing training stability (see Section \ref{sec:stability_analysis}).
    Another major distinction from prior works~\citep{watter2015embed, li2020causal, yin2022embedding, mondal2024efficient} is that we retain only LSP and CP as the essential learning principles. Prior work~\citep{ni2023bridginga} has shown that LSP and CP are sufficient to derive the cost-to-go predictor ($\phi_{V^*}$), and that the state reconstruction (SR) is unnecessary since our goal is not to accurately predict the next physical state but to plan directly in the latent space. Moreover, we maintain a quadratic cost form, unlike methods that use neural network–based cost models~\citep{mondal2024efficient}, which allows the use of LQR for efficient control.
    \textit{In summary, our objective is a minimal yet expressive implementation that learns a latent linear system for efficient control with minimal prior knowledge.}

\noindent\textbf{Real-time Predictive Control.} 
    For nonlinear systems, beyond the computationally expensive SQP method, several sampling-based predictive control approaches have been proposed for real-time deployment in robotic tasks, such as cross-entropy method (CEM)~\citep{deboer2005tutorial}, model predictive path integral (MPPI)~\citep{williams2015dataa} and predictive sampling~\citep{howell2022predictive}. These methods can achieve speedup with efficient GPU implementations. In comparison, our method requires only a nonlinear embedding function and a linear feedback control law, achieving up to a twenty-fold speedup for complex tasks on CPU-based systems, as shown in Section~\ref{sec:main_results}. 
    While recent works explore representation learning for control~\citep{zhang2024sample, lutkus2025representations, toso2025learning}, they primarily address safety or stability in the latent system. In contrast, our paper focuses on leveraging latent structure to specifically improve system efficiency.

\section{Latent Linear Quadratic Regulator}
\label{sec:latent_linear_mpc}

   We provide a detailed introduction to model predictive control and the Koopman operator in Appendix~\ref{sec:background}. 
   Building on this, we aim to mitigate the high computational burden of the SQP approach in nonlinear MPC, thereby enabling real-time control in robotic tasks. To achieve this, we transform the nonlinear dynamic model into a linear one in a latent space inspired by the Koopman operator. Unlike prior work, we adopt the \textit{Brunovsky canonical form} for latent linear dynamics and $\textit{jointly learn}$ the nonlinear embedding and quadratic cost function, allowing the efficient linear quadratic regulator (LQR) to compute optimal controls. Finally, we analyze the stability of the proposed structure.

    % % It is well known that nonlinear MPC suffers from a high computational burden and is thus hard to deploy on embedded devices. 
    % This paper tackles this problem by transforming the nonlinear MPC formulation into a latent linear one and adopting a linear quadratic regulator (LQR) to acquire the optimal control inputs. We learn the parameters of the new formulation by imitating the original nonlinear MPC. In the end, we analyze the stability of to validate the proposed structure.
    
\subsection{Latent Linear Quadratic Problem}
\label{sec:latent_linear_quadratic_problem}
    
    \begin{figure*}[ht!]
    \centering
    \subfigure[Nonlinear Model]{%
            \label{fig:nonlinear_model}% label for this sub-figure
            \includegraphics[width=0.23\linewidth]{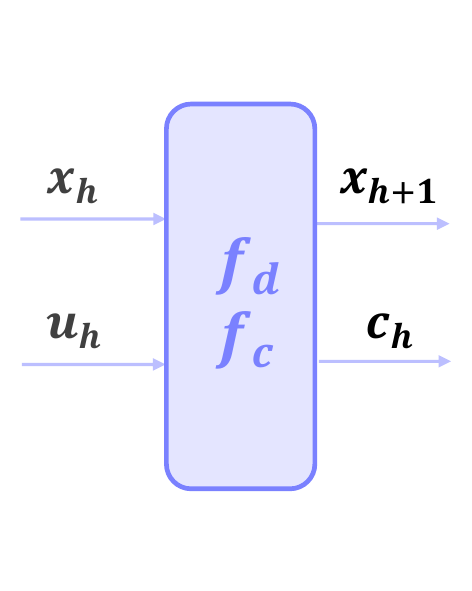}
    }
    \subfigure[Latent Linear-Quadratic Model]{%
            \label{fig:latent_linear_model}% label for this sub-figure
            \includegraphics[width=0.63\linewidth]{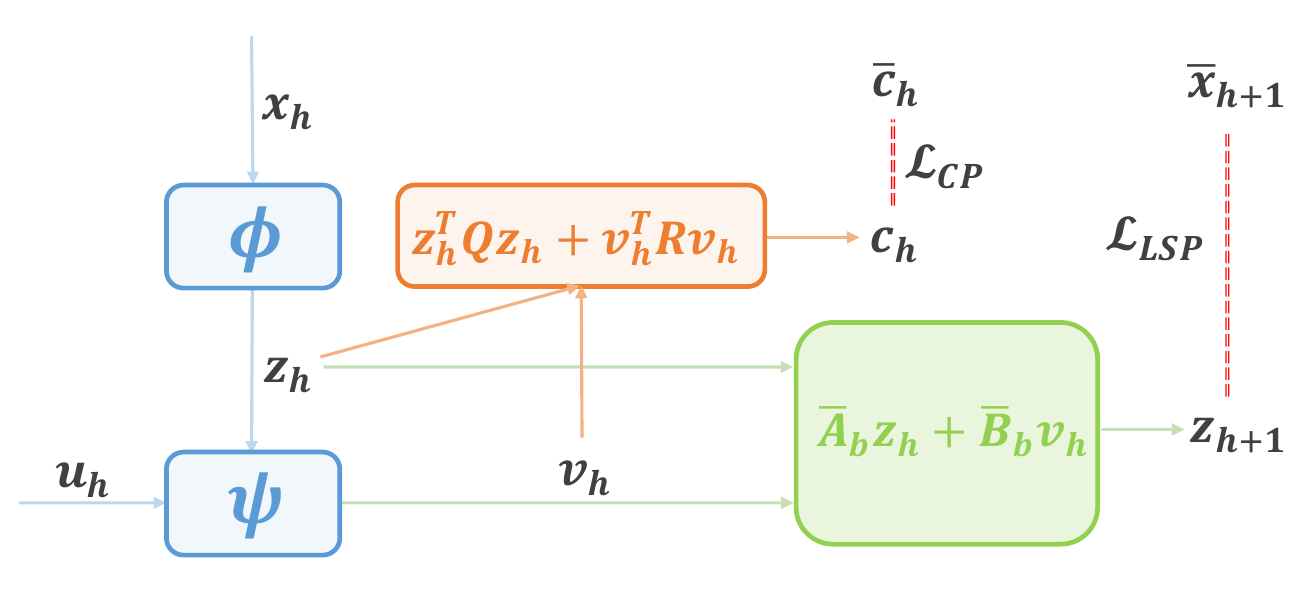}
    }
    \caption{Visualization of different model representations for the same problem. \textbf{Left:} nonlinear model with nonlinear dynamic model $f_d$ and cost function $f_c$.
    \textbf{Right:} equivalent latent linear-quadratic model with linear dynamic model, quadratic cost function and nonlinear embeddings. $\bar{c}$ and $\bar{x}$ represent ground-truth values for given training dataset.}
    \label{fig:models} % label for whole figure
    \end{figure*}
    
    In a standard nonlinear MPC setup, both the dynamic model $x_{h+1} = f_d(x_h, u_h)$ and the cost function $c_h = f_c(x_h, u_h)$ are nonlinear, with state $x_h \in \mathbb{R}^n$ and control $u_h \in \mathbb{R}^m$, as illustrated in Figure~\ref{fig:nonlinear_model}. We instead represent this nonlinear system in a latent linear one, as shown in Figure~\ref{fig:latent_linear_model}. Specifically, the state $x_h$ is first lifted to a finite-dimensional latent state $z_h \in \mathbb{R}^N, N > n$ with the embedding function $\phi: \mathbb{R}^n \to \mathbb{R}^N$. The control $u_h$ and latent state $z_h$ are combined to yield a latent control $v_h \in \mathbb{R}^m$ via another embedding $\psi: \mathbb{R}^m \times \mathbb{R}^N \rightarrow \mathbb{R}^m$. Furthermore, we assume the latent dynamics are linear: $z_{h+1} = A z_{h} + B v_h$, where $A \in \mathbb{R}^{N \times N}$ and $B \in \mathbb{R}^{N \times m}$, and the cost is quadratic: $z_h^TQz_h + v_h^TRv_h$ with positive semi-definite matrices $Q \in \mathbb{R}^{N \times N}$ and  $R \in \mathbb{R}^{m \times m}$. The resulting latent linear–quadratic problem becomes:

 %    \begin{align}
	%    \text{Equation~\ref{eq:mpc}} \quad \Longrightarrow \quad
 %        \begin{split}
	%    \underset{v_0, v_1, \cdots, v_H}{\text{min}} & \quad \sum_{h=0}^{H} z_h^TQz_h + v_h^TRv_h\\
	%    \text{s.t.} 
 %        & \quad z_{h+1} = Az_h + Bv_h, \quad z_0=\phi(x_0) = \phi(x(t)). \\
 %    % & \quad z_h = \phi(x_h) \quad F(z_h^TQz_h + u_h^TRu_h)=c(x_h, z_h).
 %    \label{eq:linear-mpc}
	% \end{split}
 %    \end{align}
    % \vspace{-0.5em}
    \begin{equation}
        \min_{v_0,\dots,v_H} \sum_{h=0}^H (z_h^\top Q z_h + v_h^\top R v_h),\ 
    \text{s.t. } z_{h+1}=Az_h+Bv_h,\ z_0=\phi(x_0)=\phi(x(t)).
    \label{eq:linear-mpc}
    \end{equation}
    This formulation is easier to solve due to its linear dynamics and quadratic cost structure. We first discuss the design of each module before presenting the control solution in Section~\ref{sec:linear_quadratic_regulator_in_latent_space}.

\noindent\textbf{Embedding Function.} 
    The embedding $\phi$ increases dimensionality ($N > n$) as suggested by the practical Koopman operator approaches and is implemented as a two-layer feedforward network. It needs not be invertible since state reconstruction is unnecessary for planning. The control embedding $\psi$ is implemented as a conditional reversible network, embedding controls as $v = \psi(u, z)$ and recovering $u$ via $u = \psi^{-1}(v, z)$. 
    % The embedding function $\phi$ is dimension-raising, since $N$ is set larger than $n$ according to Koopman operator theory. It is implemented as a 2-layer feed-forward neural network. Notably, it does not have to be reversible, as we don't need to reconstruct the next state to make plans. The control embedding function $\psi$ is implemented as a conditional reversible neural network. The forward pass is to embed the control $u$ into the latent control $v$ conditioned on the latent state $z$, i.e. $v = \psi(u, z)$. After planning on the latent space and acquiring the optimal latent control $v^*$, we can transfer it back to the original control space by $u^* = \psi^{-1}(v^*, z)$. 
    In practice, $\psi$ is represented with two learnable matrices $M \in \mathbb{R}^{m \times m}$ and $W \in \mathbb{R}^{m \times N}$ that $v = \exp(\frac{M-M^T}{2}) (u - Wz) $ and $u =  \exp(\frac{M^T-M}{2})v + Wz$.

\noindent\textbf{Dynamic Model.}
    A controllable linear time-invariant (LTI) system with $N$ states and $m$ inputs is constructed in the latent space.  
    % Various formulations exist for characterizing such systems.
    Here, we adopt the most fundamental formulation—a parametrization based on controllability indices expressed in the Brunovsky canonical form\footnote{We avoid referring to the Brunovsky canonical form as a “Jordan canonical form with zero eigenvalues,” since such a description only characterizes $\bar{A}_b$ while neglecting $\bar{B}_b$.}~\citep{brunovsky1970classification}
    % \vspace{-0.5em}
    \begin{equation}
    \begin{aligned}
        \bar{A}_b &= \text{blkdiag}(A_{i}) \quad 
        A_{i} = \begin{bmatrix}
            \mathbf{0}_{ (u_i-1) \times 1}  & \mathbf{I}_{\mu_i-1} \\
            0 & \mathbf{0}_{ 1 \times (u_i-1)}
        \end{bmatrix} \in \mathbb{R}^{\mu_i \times \mu_i}, \\
        \bar{B}_b &= \text{blkdiag}(B_{i}) \quad 
        B_i = \begin{bmatrix}
            \mathbf{0}_{ (u_i-1) \times 1}  \\
            1 \\
        \end{bmatrix} \in \mathbb{R}^{\mu_i \times 1},
    \end{aligned}
    \label{eq:canonical-form-structure}
    \end{equation}
    where $\mathbf{I}_j$ denotes identity matrix of size $j \times j$, $\mathbf{0}_{j \times k}$ denotes zero matrix of size $j \times k$, and $\{\mu_i\}_{i=1}^m$ denotes the set of controllability indices satisfying $\sum_{i=1}^m \mu_i = n$. 
    These indices are invariant under any nonsingular state transformation $T$ and input transformation $G$, i.e., if $(A, B)$ is controllable, then $(TAT^{-1}, TBG)$ has the same controllability indices~\citep{antsaklis2007linear}.
    In our framework, the linear transformations $T$ and $G$ can be absorbed into the embedding functions $\phi$ and $\psi$, allowing us to fix the most compact representation—the Brunovsky canonical form—throughout the learning process.
    Compared with randomly fixed or freely learnable $(A, B)$ pairs, using the Brunovsky canonical form $(\bar{A}_b, \bar{B}_b)$ offers several advantages: (1) All eigenvalues of $\bar{A}_b$ are zero. In contrast, random $A$ matrices may contain eigenvalues with magnitudes greater than one, leading to numerical instability when simulating the linear dynamics $x^+ = Ax + Bu$ involving matrix powers of $A$. (2) Solving the LQR problem with $(\bar{A}_b, \bar{B}_b)$ is numerically more efficient~\citep{yang2025brunovsky}.
    To the best of the author's knowledge, there was no prior work providing principled guidance on selecting controllability indices of the linear latent system.
    Nonetheless, when all indices are identical, the resulting deadbeat controller $K$ (which places all eigenvalues of $A - BK$ at zero) is unique~\citep{schlegel1982parameterization}.
    Consequently, in practice, we set $\mu_i = \lfloor N / m \rfloor, \forall i$ to constrain the solution space and promote training stability.
    There is another line of work~\citep{lusch2018deep, mondal2024efficient} using the diagonal structure in the matrix $A$, but they are either complex-valued or uncontrollable in the latent space. We will analyze the differences from other structures theoretically and empirically in Section~\ref{sec:stability_analysis}.
    % We assume the latent linear model $(A, B)$ is controllable, ensuring the existence of a stabilizing feedback policy~\citep{antsaklis2007linear}. 
    % According to \citet{brunovsky1970classification}, any controllable pair $(A, B)$ can be transformed into the Brunovsky canonical form $(\bar{A}_b, \bar{B}_b)$ (Eq.~\ref{eq:canonical-form-structure}), characterized by controllability indices $\{\mu_i\}_{i=1}^m$.
    % Since we have left sufficient degrees of freedom to the embedding functions $\phi$ and $\psi$, we fix the linear dynamic model to be the Brunovsky canonical form without loss of generality. Thus, they can be viewed as fixed matrices that do not require learning. Following \citet{amin1988parameterization}, when all controllability indices are equal, a unique linear feedback controller exists that stabilizes the system with all eigenvalues at zero. Thus, we empirically set the controllability indices to be constant $\mu_i = \lfloor N/m \rfloor$, to limit the solution space and enhance training stability.
    
\noindent\textbf{Cost Function.}
    $Q \in \mathbb{R}^{N \times N}, R \in \mathbb{R}^{m \times m}$ are positive semi-definite matrices for reasonable control performances. To achieve this, we introduce two auxiliary lower triangular matrices $L_Q \in \mathbb{R}^{N \times N}, L_R \in \mathbb{R}^{m \times m}$ and construct $Q=\mathbf{I}_{N}+ L_Q L_Q^T$ and $R=\mathbf{I}_{m} + L_R L_R^T$. 

\subsection{Linear Quadratic Regulator in Latent Space}
\label{sec:linear_quadratic_regulator_in_latent_space}
    
    Given the latent linear system $(\bar{A}_b, \bar{B}_b)$ and quadratic cost $(Q, R)$, the infinite-horizon LQR provides an efficient control law. We extend the horizon $H$ in Equation~\ref{eq:linear-mpc} to $\infty$ for practical reasons: (1) the practical value $H$ in MPC is usually large enough and its solution is equivalent to the infinite horizon; (2) it is easy to design an efficient control law with infinite horizon~\citep{rawlings2020model}. The optimal gain matrix $K \in \mathbb{R}^{m \times N}$ is obtained by solving the discrete-time algebraic Riccati equation. Due to the Brunovsky-form dynamics and the positive semi-definite cost matrices, the Riccati solution is numerically stable (see Section~\ref{sec:stability_analysis}). In the end, the optimal control in the latent space is calculated by $v^* = - Kz$, and the optimal control in the original action space is recovered by the invertible embedding function as $u^* = \psi^{-1}(v^*, z)$. As the gain matrix $K$ is pre-computed offline, the only online computation per step is calculating $u^* = \psi^{-1}(v^*, z) = \psi^{-1}(-Kz, z) = \psi^{-1}(-K \phi(x), \phi(x))$, given current state $x$.

\subsection{Parameters Identification}
\label{sec:parameters_identification}

    Before applying LaLQR during online deployment, we identify the parameters $(\phi, \psi, Q, R)$ in a data-driven manner, keeping $(\bar{A}_b, \bar{B}_b)$ fixed. As discussed in Section~\ref{sec:related_work}, we retain only two learning objectives: latent state prediction (LSP) and cost prediction (CP), for simplicity and expressiveness. 
    Let $(x_h, u_h, \bar{c}_h, \bar{x}_{h+1})$ denote the ground-truth values directly obtained from the training dataset.

    \noindent\textbf{Latent State Prediction (LSP)} enforces consistency of the latent dynamics across consecutive steps.
    Given a sampled transition tuple $(x_h, u_h, \bar{x}_{h+1})$ satisfying the nonlinear dynamic model $\bar{x}_{h+1} = f_d(x_h, u_h)$, we define the latent state prediction loss as $ \mathcal{L}_{LSP}(x_h, u_h, \bar{x}_{h+1};\phi, \psi) = \| \text{sg}(\bar{z}_{h+1}) - (\bar{A}_b z_h + \bar{B}_b v_h)\|_2^2$, where $\bar{z}_{h+1} = \phi(\bar{x}_{h+1}), z_h = \phi(x_h), v_h = \psi(u_h, z_h)$ and $\text{sg}$ denotes the stop-gradient operator to prevent representation collapse (no gradients passed to 
    $\bar{z}_{h+1}$).
    
    \noindent\textbf{Cost Prediction (CP)} predicts the cost in the latent space, a crucial but often overlooked component for deriving optimal controllers. Additionally, this objective acts as an implicit regularizer on the latent representation, encouraging it to preserve task-relevant features that are important for control.
    Given a sampled tuple $(x_h, u_h, \bar{c}_{h})$ satisfying the nonlinear cost function $\bar{c}_h = f_c(x_h, u_h)$, we align it with the latent quadratic cost $z_h^TQz_h+v_h^TRv_h$. Since the cost function $f_c$ can be arbitrarily nonlinear, a direct equivalence with a quadratic cost might be intractable. 
    We assume either that the true form of the cost function is known or that only sampled costs $\bar{c}_h$ are accessible when the cost is computed externally, e.g., in multi-armed bandits~\citep{sutton1998reinforcement}.
    Instead, we introduce a monotonic function $F$ to match the scales of two costs, and thus the cost prediction loss is defined as $ \mathcal{L}_{CP}(x_h, u_h, \bar{c}_h;\phi, \psi, Q, R, F) = \| \bar{c}_h - F(z_h^TQz_h+v_h^TRv_h)\|_2^2$, where $z_h = \phi(x_h), v_h = \psi(u_h, z_h)$. In practice, we implement $F$ as the Lipschitz monotonic networks (LMN)~\citep{nolte2022expressive}, allowing the monotonic function itself to be also trainable. 
    LMN is an additional component applied on top of the cost function and does not affect the linearity of the latent system.

    These two objectives can be seen as different levels of abstraction of the original states $x$. Prior work~\citep{ni2023bridginga} has shown that LSP and CP are sufficient to derive the abstraction level of the cost-to-go predictor ($\phi_{V^*}$). Besides, \citet{ni2023bridginga} observes that the state reconstruction (SR) abstraction is easily influenced by noisy states and outliers, and is unnecessary for planning. The above explains why we employ only the LSP and CP principles for parameter identification. 
    The parameters $(\phi, \psi, Q, R, F)$ are jointly learned by minimizing a combined objective of LSP and CP: $\min_{(\phi, \psi, Q, R, F)} \frac{1}{|\mathcal{D}|} \sum_{(x_h, u_h, \bar{x}_{h+1}, \bar{c}_h) \in \mathcal{D}}  \mathcal{L}_{LSP}(x_h, u_h, \bar{x}_{h+1};\phi, \psi)  +  \mathcal{L}_{CP}(x_h, u_h, \bar{c}_h;\phi, \psi, Q, R, F)$, where $\mathcal{D}$ denotes the training dataset containing multiple transitions. The approximation errors can be bounded by constraining the magnitudes of all learnable components.
    
\subsection{Stability Analysis}
\label{sec:stability_analysis}

    %%%%%%%%%%%%%%%%%%%%%%%
    % % fig: MPC dynamics
    \begin{figure}[htbp]
    \centering
        \subfigure[Control Rank]{%
            \label{fig:control_rank}% label for this sub-figure
            \includegraphics[width=0.3\linewidth]{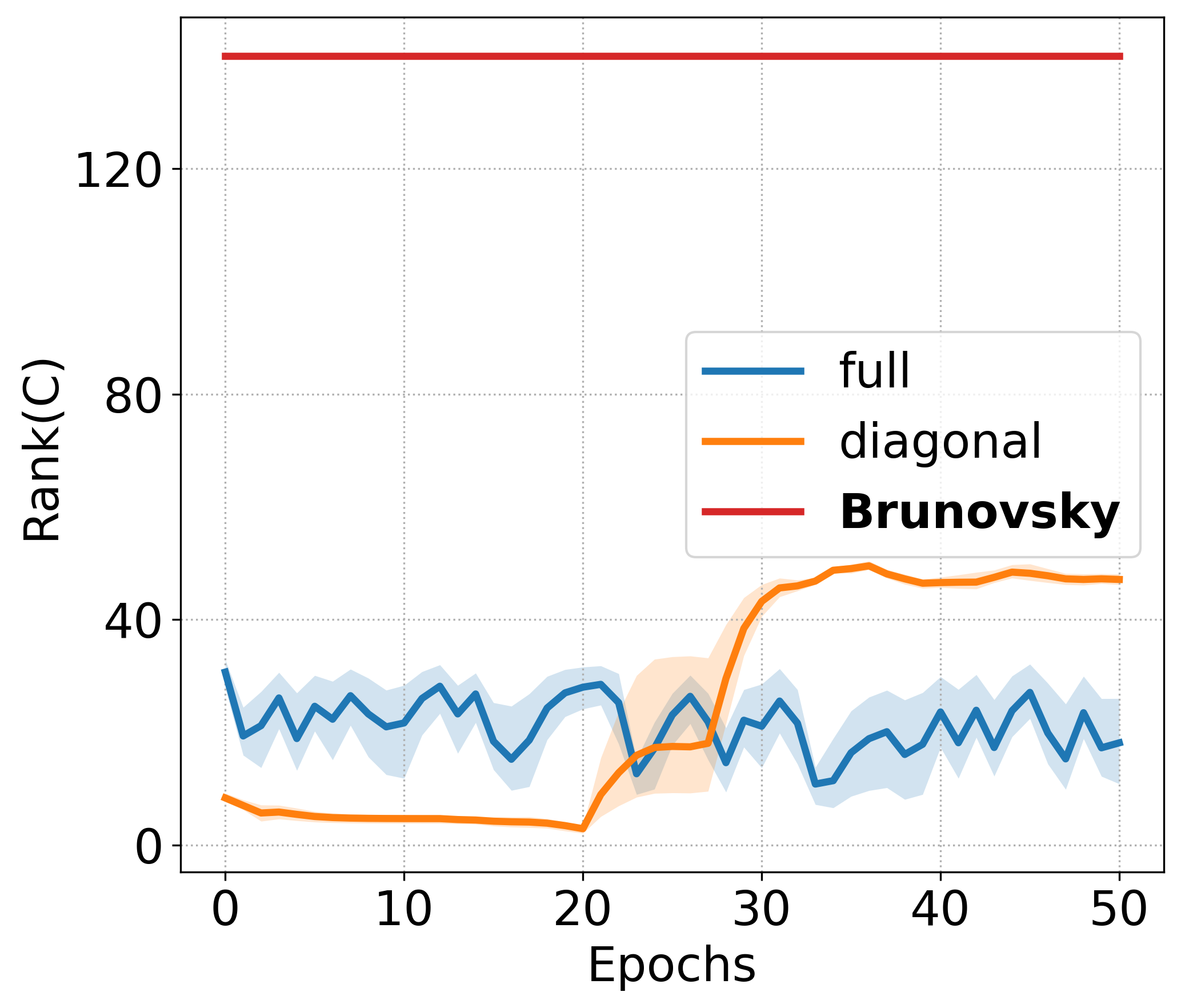}
        }
        \subfigure[Eigenvalues]{%
            \label{fig:eigenvalues}% label for this sub-figure
            \includegraphics[width=0.3\linewidth]{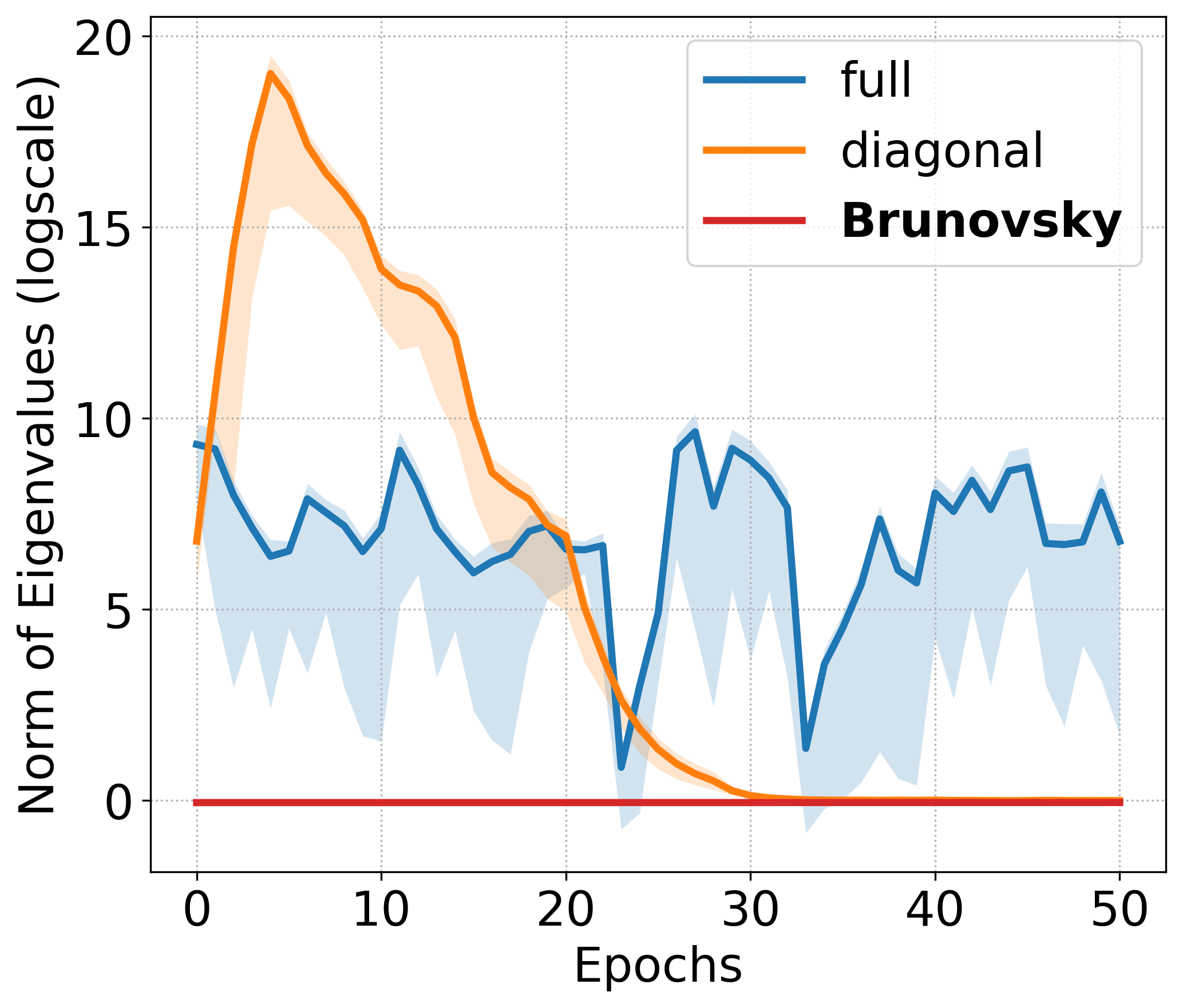}
        }
        \subfigure[Eigenloss]{%
            \label{fig:eigenloss}% label for this sub-figure
            \includegraphics[width=0.3\linewidth]{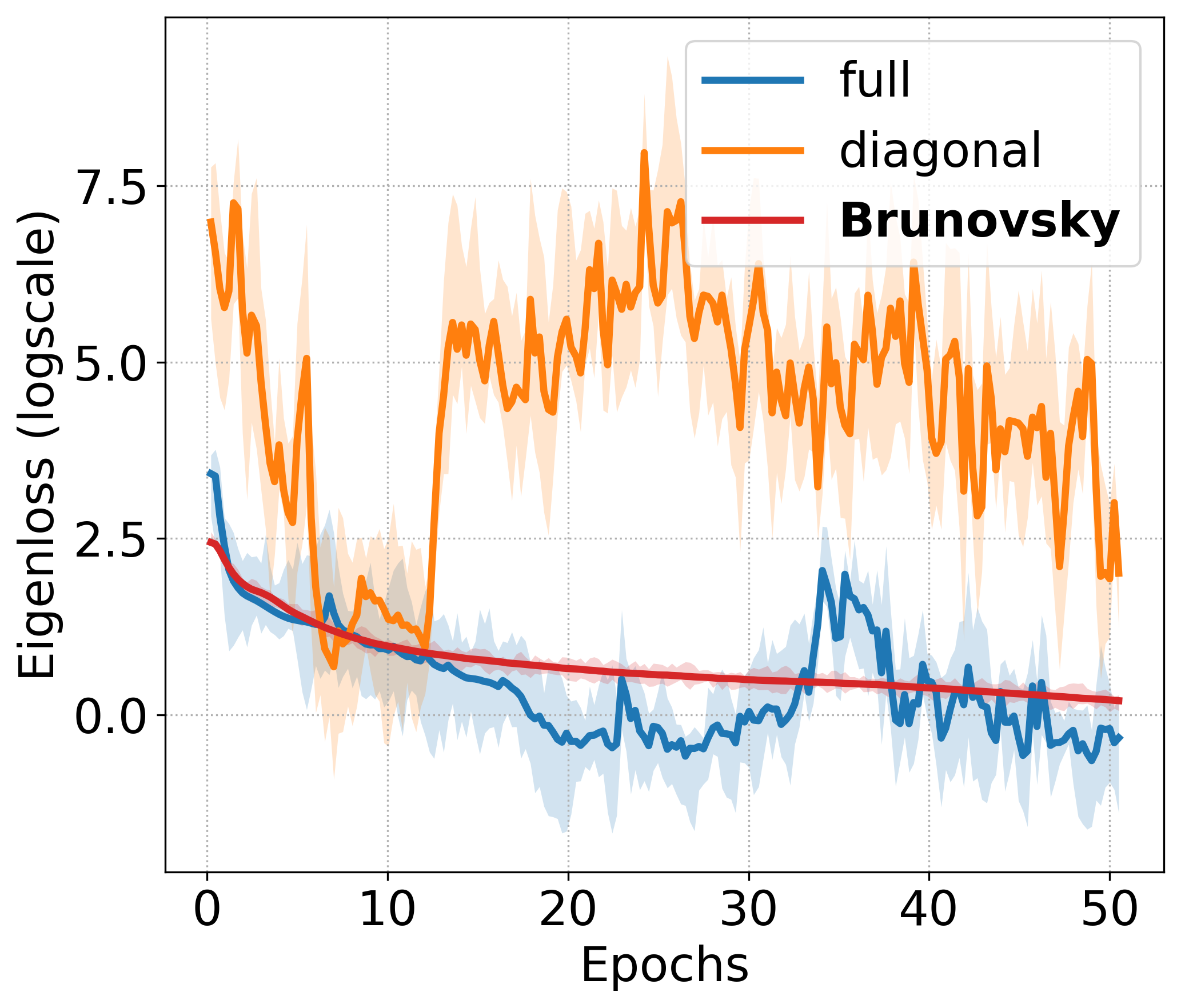}
        }
    \caption{Stability analysis on the dynamic of LaLQR on \texttt{Panda} tasks. Each curve shows the average values over three seeds to follow previous work~\citep{hansen2022temporal}, with shaded regions denoting the standard deviation.}
    \label{fig:stability_analysis} % label for whole figure
    \end{figure}
    % %%%%%%%%%%%%%%%%%%%%%%%
    
    In this section, we evaluate the stability induced by using the Brunovsky canonical form in the latent dynamics. From Equation~\ref{eq:linear-mpc}, different feedback laws $K$ arise from different $(A, B)$ structures, and the closed-loop matrix $L=A - BK$ governs latent-space stability in the latent space $z$. We compare our choice of Brunovsky canonical form $(\bar{A}_b, \bar{B}_b)$ to two additional baselines: "full" means the full matrix $A$ and $B$ are learnable, 'diagonal' means $A$ is a learnable diagonal matrix and each dimension of $z$ is independent. All parameters are identified following Section~\ref{sec:parameters_identification}.
    For each $(A,B,K)$, we evaluate the controllability via the rank of the controllability matrix $C = [B, AB, \dots A^{n-1}B]$, and stability via the mean absolute value of the eigenvalues of the closed-loop system matrix $L=A-BK$. 
    We also compare the learned latent system with the original nonlinear system at the stable point $x_S,u_S$. Specifically, the nonlinear system can be linearized around the stable point to yield a closed-loop system: $x_{h+1} =  L^o x_h$, where the matrix $L^o=\frac{\partial f_d}{\partial x}\big|_{x=x_S}   - \frac{\partial f_d}{\partial u}\big|_{u=u_S} K^o $ with the feedback gain matrix $K^o$.
    Meanwhile, for the latent system with closed-loop matrix $L=A-BK$ and Jacobian $J = \frac{\partial \phi}{\partial x}|_{x_S}$, the corresponding latent closed-loop system is $J x_{h+1} = L J x_h$. The latent closed-loop system $J x_{h+1} = LJx_h$ accurately represents the linearized closed-loop system $x_{h+1} = L^o x_h$ if, \textit{for every eigenvalue–eigenvector pair $(\lambda, v)$ of $L^o$, $(\lambda, Jv)$ is also an eigenpair of $L$}. Accordingly, we define $\textit{eigenloss}$ as $ \sum_i \|LJv_i- \lambda_i Jv_i\|_2^2$ computed over all eigenvalue-eigenvector pairs of matrix $L^o$.
  
   % \begin{proposition}[Eigenstructure Equivalence]
   % \label{prop:eigenvalues_and_eigenvectors_of_equivalent_systems}
   %      The latent closed-loop system $J x_{h+1} = LJx_h$ is locally equivalent to the linearized closed-loop system $x_{h+1} = L^o x_h$ if, for every eigenvalue–eigenvector pair $(\lambda, v)$ of $L^o$, the pair $(\lambda, Jv)$ is also an eigenpair of $L$.
   %  \end{proposition}

    We visualize the above-mentioned metrics in Figure~\ref{fig:stability_analysis}. As shown in Figure~\ref{fig:control_rank}, our Brunovsky-based model maintains full controllability throughout training, whereas the alternatives exhibit degraded rank in the controllability matrix $C$ with the variations of $(A,B)$. 
    For the full matrix with maximum representational capacity, its effective rank may still be low due to approximation errors during learning.
    Similarly, the eigenvalue magnitudes (Figure~\ref{fig:eigenvalues}) remain small for the Brunovsky canonical form, confirming the stable closed-loop autonomous system. Finally, Figure~\ref{fig:eigenloss} shows that our model most accurately approximates the original system, while the diagonal structure lacks sufficient expressivity.

\section{Experiments}
\label{sec:experiments}

    We empirically evaluate LaLQR to verify its ability to learn a latent linear–quadratic structure that enables efficient and stable control across diverse robotic systems. 
    
    %%%%%%%%%%%%%%%%%%%%%%%
    % % fig: MPC dynamics
    \begin{figure}[htbp]
    \centering
        \subfigure[$\texttt{Cartpole}$]{%
            \label{fig:cartpole}% label for this sub-figure
            \includegraphics[width=0.23\linewidth]{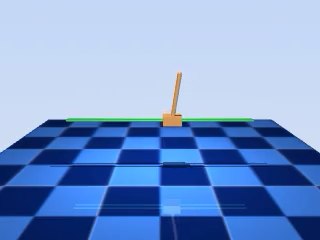}
        }
        \subfigure[$\texttt{Particle}$]{%
            \label{fig:swimmer}% label for this sub-figure
            \includegraphics[width=0.23\linewidth]{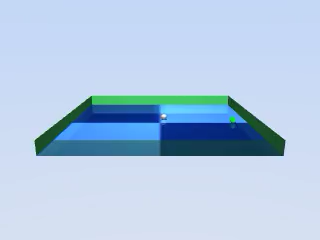}
        }
        \subfigure[$\texttt{Swimmer}$]{%
            \label{fig:quadrotorl}% label for this sub-figure
            \includegraphics[width=0.23\linewidth]{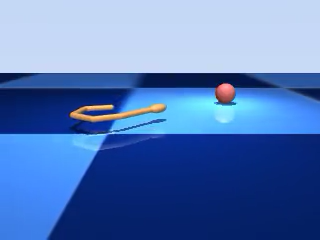}
        }
        \subfigure[$\texttt{Panda}$]{%
            \label{fig:panda}% label for this sub-figure
            \includegraphics[width=0.23\linewidth]{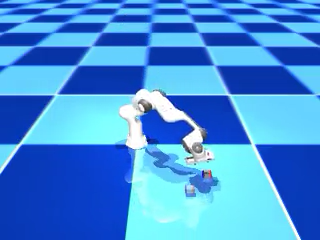}
        }
    \caption{Visualization of robots used in the experiments, with increasing complexity in action dimension $m \in \{1, 2, 5, 7\}$ and state dimension $n \in \{4, 4, 16, 33\}$.}  
    % \vspace{-2em}
    \label{fig:robots} % label for whole figure
    \end{figure}
    % %%%%%%%%%%%%%%%%%%%%%%%
   
%===============================================================================
\subsection{Experimental Setup}
\label{sec:experimental_setup}

\noindent\textbf{Task Setup.}
    We use MuJoCo~\citep{todorov2012mujoco} as the simulation platform to evaluate all baselines. Four robotic tasks are selected, $\texttt{Cartpole}$, $\texttt{Particle}$, $\texttt{Swimmer}$, and $\texttt{Panda}$, with increasing complexity in action and state spaces, as illustrated in Figure~\ref{fig:robots}. For \texttt{Cartpole}, the controller must balance the pole at the origin. For \texttt{Particle} and \texttt{Swimmer}, the goal is to reach a specified target position. For \texttt{Panda}, the task is to control the robotic arm to pick and place a cube.
    Detailed task descriptions are provided in Appendix~\ref{sec:robotic_tasks_on_mujoco}. Training data are collected by rolling out expert controllers (SQP) for 200 episodes, each with a horizon of 500 steps. The use of SQP can be extended to other efficient—or even imperfect—controllers to facilitate scalable data generation; we provide a detailed analysis of this extension in Section~\ref{sec:generalization}.

\noindent\textbf{Baselines.} 
    We compare our proposed method, \textbf{LaLQR}, against the following baselines:
    \textbf{(1) Sequential quadratic programming (SQP)} adopts the simulator's ground-truth nonlinear dynamic model of $x_{h+1} = f_d(x_h, u_h)$ and cost function $f_c(x_h, u_h)$, with the sequential quadratic problem (SQP) method~\citep{tassa2012synthesis} to generate optimal controls.
    \textbf{(2) Cross-entropy method (CEM)} is a sampling-based MPC method to solve Equation~\ref{eq:mpc}.
    \textbf{(3) Local LQR (LoLQR)} expands the nonlinear dynamic model at the stable point $(x_S, u_S)$ using a first-order Taylor expansion and applies the resulting linear model globally. Note that identifying an accurate stable point is challenging for some systems, such as \texttt{Swimmer}.
    \textbf{(4) Imitation learning (IL)} learns a direct control policy $u_h^* = \pi(x_h)$ using a 3-layer feedforward neural network with a comparable number of parameters to LaLQR for fairness. IL learns to imitate the feedback control law directly; thus, its performance depends heavily on expert data quality and typically generalizes poorly—an issue analyzed further in Section~\ref{sec:generalization}.
    Planning hyperparameters for SQP, CEM, and LoLQR, as well as training hyperparameters for IL and LaLQR, are listed in Appendix~\ref{sec:hyperparameters_of_baselines}.
% $\textbf{- LaLQR (Ours)}$ represents the introduced method in Section~\ref{sec:latent_linear_mpc}. After imitation learning the nonlinear MPC as Algorithm~\ref{alg:MPC_imitation_learning}, the gain matrix $K$ is further calculated, and the final control law is set as $u_h^* = \psi^{-1}(- K\phi(x_h), \phi(x_h))$. 
     
\subsection{Main Results}
\label{sec:main_results}

     % ------------------------
    % all figures
    \begin{figure*}[ht]
        \centering
        \includegraphics[width=1.0\linewidth]{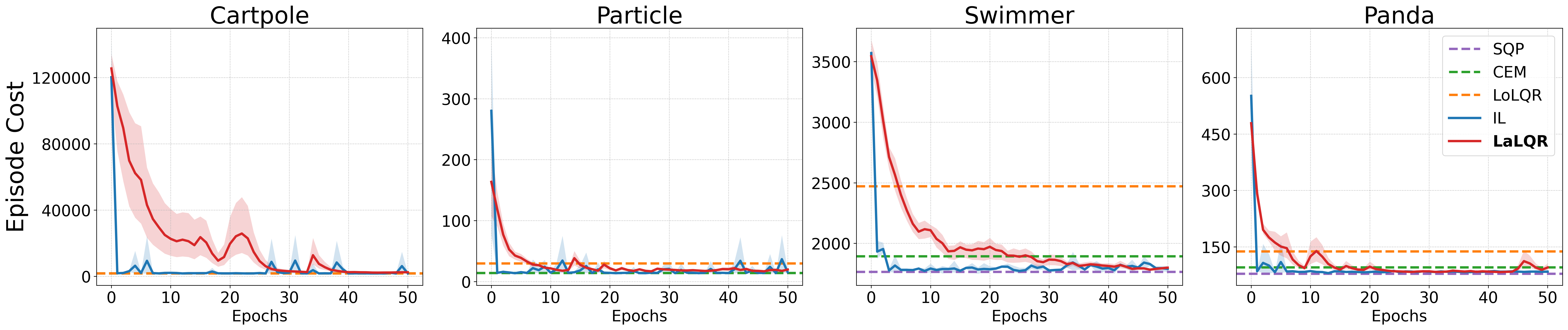}
        \caption{Training curves of all baselines on MuJoCo tasks. Each curve shows the average episode cost over three seeds, with shaded regions denoting the standard deviation.}
        \label{fig:training_results}
    \end{figure*}
    
    % method comparison
    \begin{table*}[ht]
    % \begin{wraptable}{r}{0.5\linewidth}
    	\begin{center}
         \scalebox{0.7}{
    		\begin{tabular}{lccccc}
            \toprule
    			\textbf{Tasks} &  \textbf{SQP} & \textbf{CEM} &  \textbf{LoLQR} & \textbf{IL} & \textbf{LaLQR}\\
            \midrule	
                \texttt{Cartpole}  & $17.64 \pm 7.21$ &  $4.64 \pm 3.03$ & $0.07 \pm 0.04$ & $0.30 \pm 0.10$ &  $0.29 \pm 0.08$  \\
            \midrule	
                \texttt{Particle} &  $20.53 \pm 1.41$ &  $4.73 \pm 2.53$ & $0.09 \pm 0.02$ & $0.28 \pm 0.04$ &  $0.31 \pm 0.03$ \\
            \midrule	
                \texttt{Swimmer} & $47.56 \pm 8.86$ & $17.55 \pm 4.50$ & $0.08 \pm 0.03$ & $4.61 \pm 2.85$ &  $4.56 \pm 1.95$ \\
             
            \midrule	
            
                \texttt{Panda} & $125.49 \pm 21.32$ & $37.67 \pm 6.78$& $0.09 \pm 0.03$ & $5.76 \pm 2.69$ & $5.72 \pm 4.61$ \\
            \bottomrule
                \end{tabular}}
    	\end{center}
    \caption{The average time (in milliseconds) for each planning step. All experiments are conducted on an x86\_64 CPU with 40 cores, and reported in average and standard deviation over 100 runs.}
    \label{tab:main_results}
    % \end{wraptable}
    \end{table*}
    
    For each task, IL and LaLQR are trained for 50 epochs using the same expert dataset. At the end of each epoch, both controllers are evaluated over 10 rollouts, and we report the average episode cost (total accumulated cost per episode) in Figure~\ref{fig:training_results}. Since SQP, CEM, and LoLQR are non-learning baselines, their results are shown as horizontal reference lines. Additional training losses are provided in Appendix~\ref{sec:additional_training_results}. 
    % rollout videos in the supplementary material.
    Figure~\ref{fig:training_results} shows that both IL and LaLQR converge toward the performance of the SQP expert, demonstrating that learning-based feedback controllers can effectively approximate optimal control laws. IL converges faster when trained on perfect expert data, since it imitates only the control outputs rather than jointly learning the embedding and cost functions as in LaLQR. However, as we show in the next section, IL's learning scheme fails to generalize to unseen conditions.
    Sampling-based CEM and locally linear LoLQR fail to match SQP on complex tasks such as \texttt{Swimmer} and \texttt{Panda}.
    Table~\ref{tab:main_results} summarizes the average computation time per planning step over 100 runs on an x86\_64 CPU with 40 cores. Although LoLQR achieves the fastest inference due to its globally linear structure, it performs the worst in control accuracy. CEM improves planning speed over SQP but remains slower than LQR-based methods. Both IL and LaLQR achieve near–real-time inference with only lightweight matrix operations, offering more than an order-of-magnitude speedup over SQP.
    In summary, LaLQR achieves a favorable trade-off between computational efficiency and control performance.
    
\subsection{Generalization}
\label{sec:generalization}

    % all figures
    \begin{figure*}[ht]
        \centering
        \includegraphics[width=0.9\linewidth]{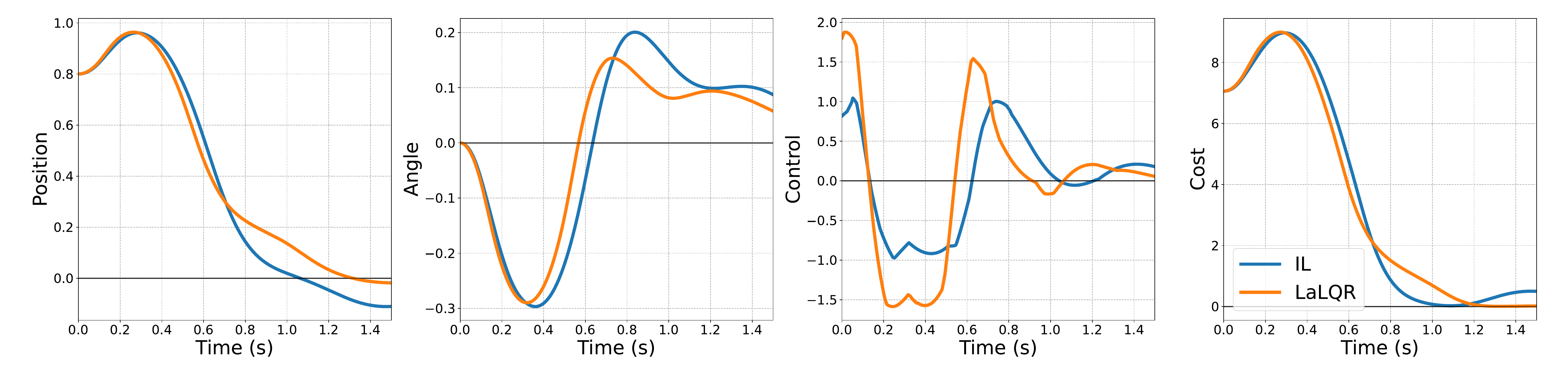}
        \caption{Control process of methods learned from imperfect experts. The x-axis is the real testing time, and the y-axis represents the partial state, control and cost in $\texttt{Cartpole}$ task.}
        \label{fig:imperfect_expert}
    \end{figure*}
    \begin{figure*}[ht]
        \centering
        \includegraphics[width=0.9\linewidth]{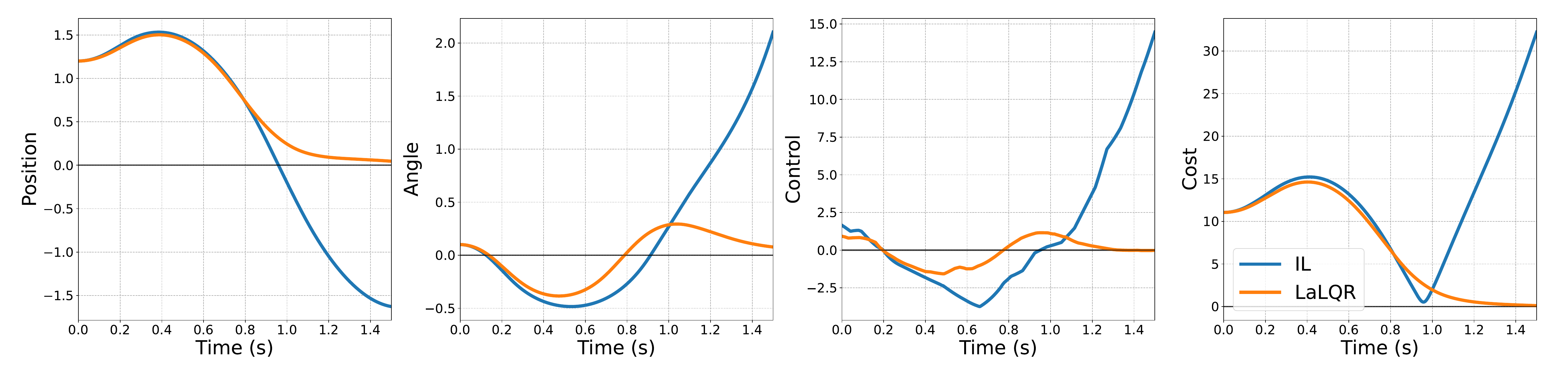}
        \caption{Control process of methods starting from unseen initial states in training. The x-axis is the real testing time, and the y-axis represents the partial state, control and cost in $\texttt{Cartpole}$ task.}
        \label{fig:initial_state}
    \end{figure*}

    Learning-based controllers often exhibit poor generalization—they perform well in training-like settings but degrade sharply under distribution shifts. We evaluate IL and LaLQR under two such scenarios: (1) imperfect expert demonstrations and (2) unseen initial states.

\noindent\textbf{Imperfect Expert Demonstrations.}
    In the previous section, both IL and LaLQR were trained on data generated by an optimal expert. Here, we introduce noise to the expert’s actions to simulate imperfect supervision: with probability $0.5$, a uniform noise sampled from $[-1, 1]$ is added to the expert control at each step. All other setups remain unchanged.
    As shown in Figure~\ref{fig:imperfect_expert}, IL directly inherits the expert’s suboptimal behavior, resulting in higher final costs. In contrast, LaLQR generalizes beyond the noisy demonstrations and still achieves stable control with near-zero final cost.
    
\noindent\textbf{Unseen Initial States.}
    We further test generalization by evaluating from unseen initial conditions. During training on the \texttt{Cartpole} task, initial positions are sampled uniformly from $[-1, 1]$ with a fixed angle of $0$. During testing, we set the initial position to $1.2$ and the angle to $0.1$. As shown in Figure~\ref{fig:initial_state}, IL fails to stabilize the system from this unseen state, while LaLQR maintains robust performance. These experiments demonstrate that learning the full MPC structure—rather than imitating actions alone—significantly enhances generalization.
    
% \vspace{-0.5em}
\subsection{Ablation Study}
\label{sec:ablation_study}

    %%%%%%%%%%%%%%%%%%%%%%%
    % % fig: MPC dynamics
    \begin{figure}[htbp]
    \centering
        \subfigure[dynamic model]{%
            \label{fig:dyn_result}% label for this sub-figure
            \includegraphics[width=0.23\linewidth]{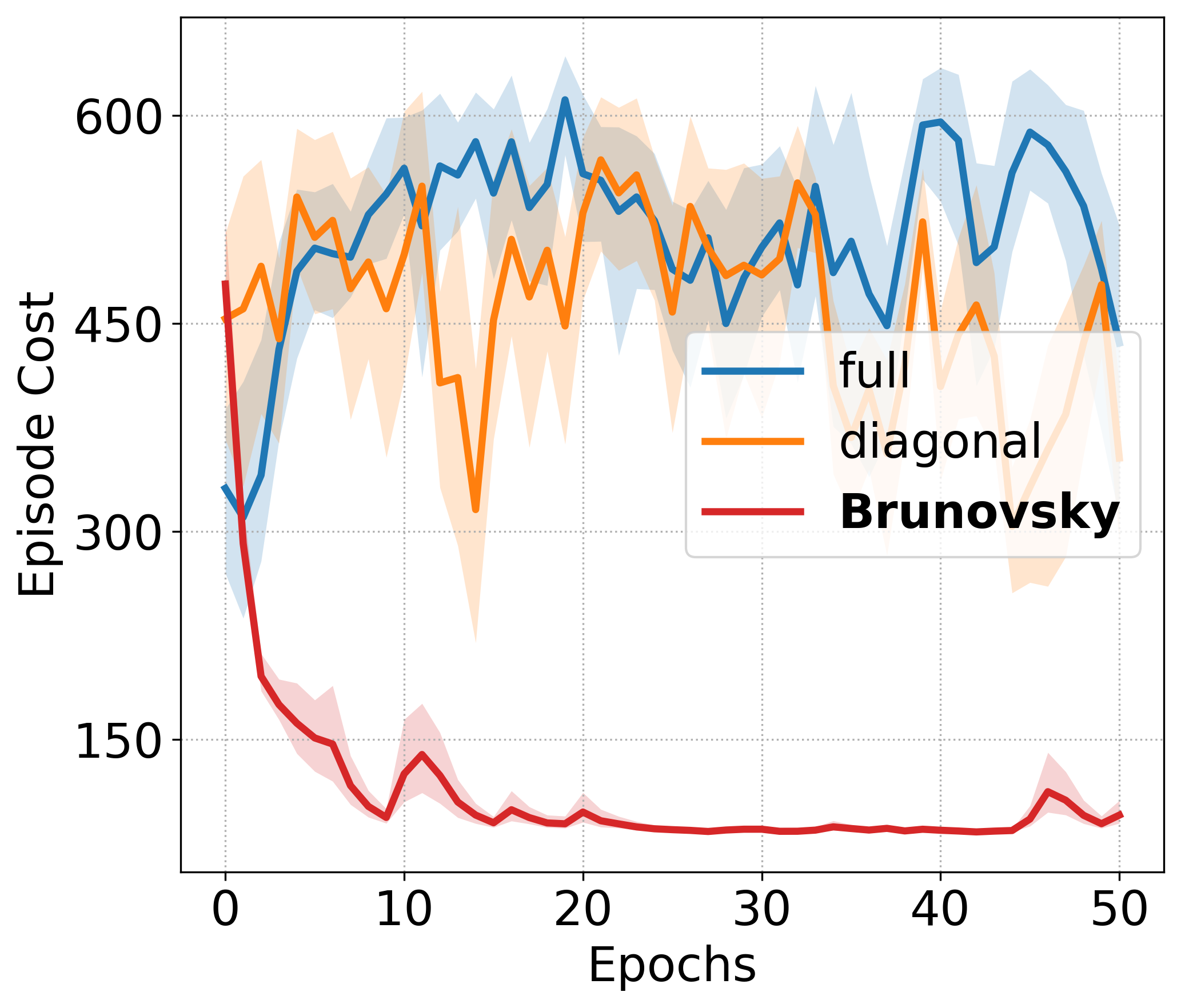}
        }
        \subfigure[latent dim]{%
            \label{fig:latent_dim_result}% label for this sub-figure
            \includegraphics[width=0.23\linewidth]{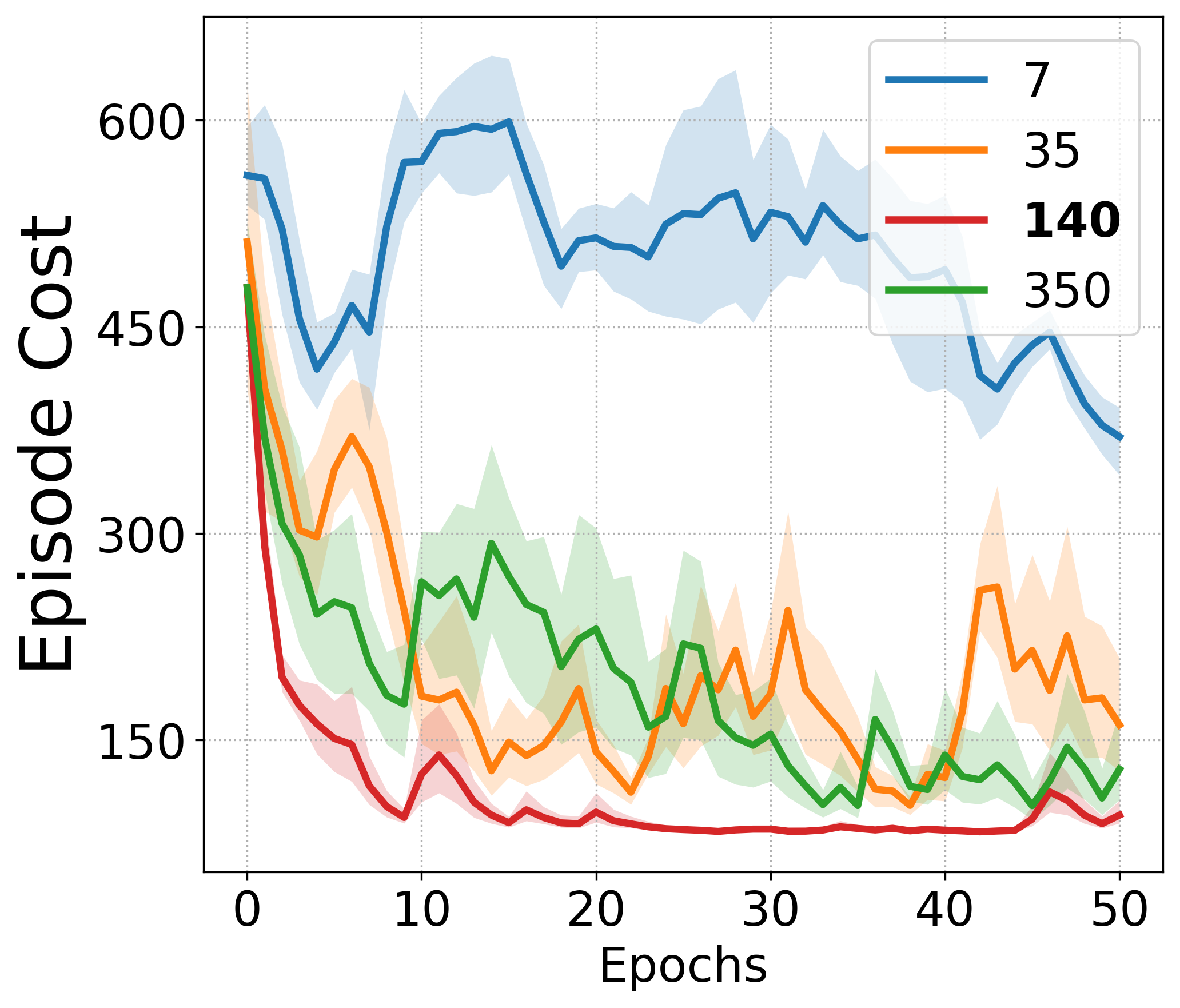}
        }
        \subfigure[cost function]{%
            \label{fig:cost_result}% label for this sub-figure
            \includegraphics[width=0.23\linewidth]{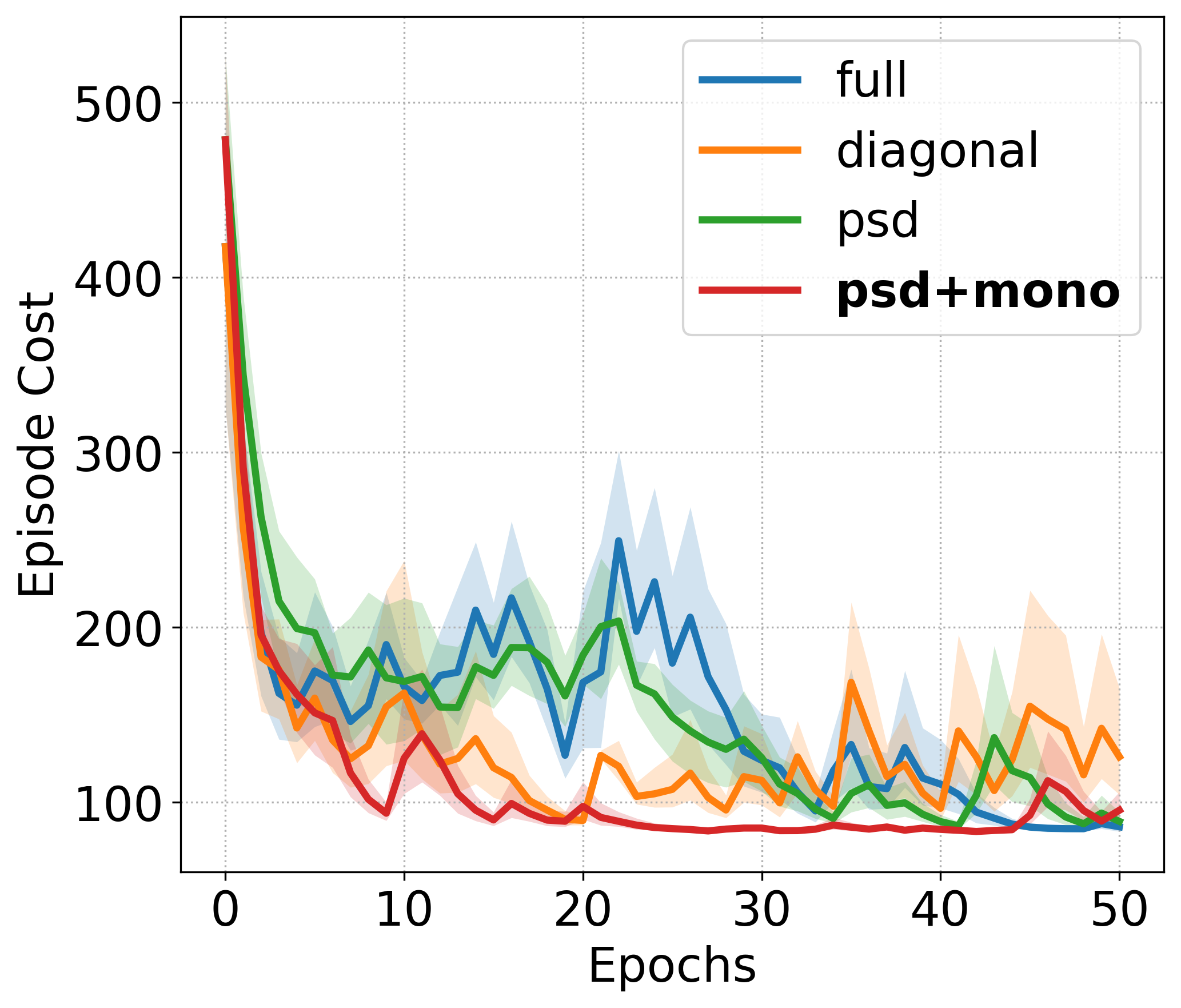}
        }
        \subfigure[cost coefficient]{%
            \label{fig:c_cost_result}% label for this sub-figure
            \includegraphics[width=0.23\linewidth]{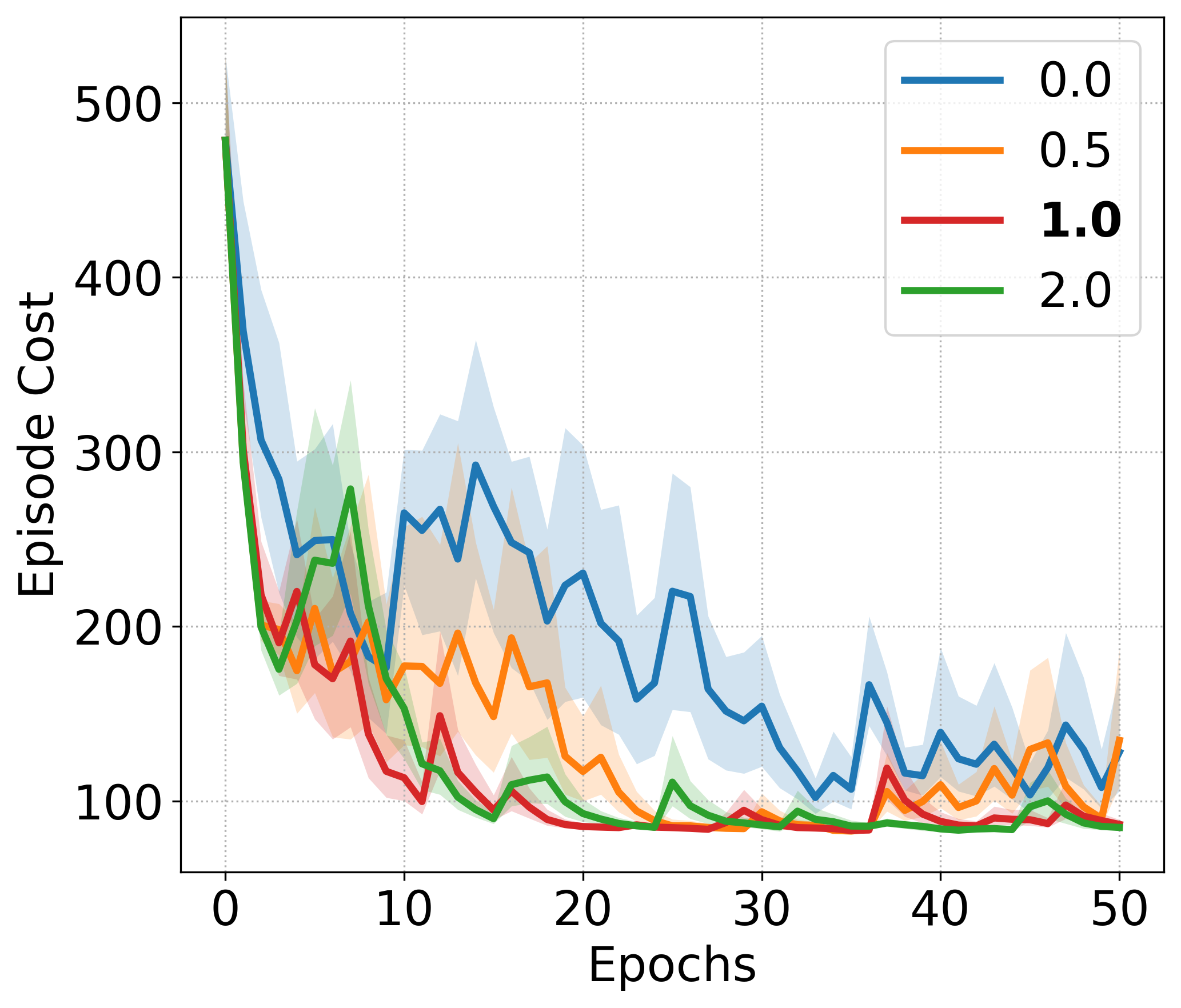}
        }
    \caption{Ablation study on key design choices of LaLQR on \texttt{Panda} tasks. Each curve shows the average episode return over three seeds, with shaded regions denoting the standard deviation.}  
    \label{fig:ablation_study} % label for whole figure
    \end{figure}
    % %%%%%%%%%%%%%%%%%%%%%%%
    
    We conduct an ablation study on the most challenging task, \texttt{Panda}, to analyze the effect of key design choices.
    First, we investigate the structure of $(A,B)$ in the latent dynamics. As shown in Figure~\ref{fig:dyn_result}, the Brunovsky canonical form $(\bar{A}_b, \bar{B}_b)$ yields both higher control performance and more stable training than the diagonal ("diagonal") or unconstrained ("full") variants, aligning with our stability analysis in Section~\ref{sec:stability_analysis}.
    Figure~\ref{fig:latent_dim_result} studies the lifted dimension $N$ of the Koopman operator. Smaller values (e.g., 7 or 35) lack sufficient expressiveness to model nonlinear dynamics, while overly large values (e.g., 350) degrade control efficiency due to large matrix computations in LQR. We thus set $N=20m$ in all experiments.
    Regarding the cost structure (Figure~\ref{fig:cost_result}), both positive semi-definite (PSD) and unconstrained $(Q,R)$ matrices achieve good performance, whereas real-diagonal $(Q,R)$ fail to capture the full eigenstructure of the nonlinear system. The monotonic function $F$ introduced in Section~\ref{sec:parameters_identification} accelerates learning by absorbing the nonlinearity between true and predicted costs. 
    Finally, Figure~\ref{fig:c_cost_result} shows the effect of the cost prediction loss coefficient, which acts as a regularization strength to prevent latent space collapse. Performance remains robust across a wide range of values, and we fix this coefficient to $1$ throughout our experiments.
   
% \subsection{Real-robot Results}
% \label{sec:real-robot_results}

% === 0.3 pages ===
\section{Conclusion}
\label{sec:conclusion}

    In conclusion, the proposed latent linear quadratic regulator (LaLQR) method effectively addresses the computational challenges of model predictive control (MPC) in systems with nonlinear dynamics. Compared to other efficient solutions, LaLQR presents better control performances and enhanced generalization capabilities. This novel approach holds promise for advancing the practical application of MPC in embedded systems, which is crucial for real-time control in complex robotic environments. 
    % Notably, there is still a performance gap against the original MPC approach especially for complex robotic tasks, which drives us to study the losses during the transformation in future work. 
    
    For future work, incorporating the embedding's estimation error with the latent LQR robustly is essential for obtaining closed-loop certificates in the original state space.
    Additionally, our method can be compared with model-based reinforcement learning algorithms~\citep{hansen2022temporal}, which simultaneously learn models and perform optimal control.
    To enhance real-world transfer, learnable modules can be combined with domain randomization~\citep{peng2018simtoreal} techniques to improve the robustness of the MPC framework. 
    Finally, instead of setting the controllability indices of the linear latent system heuristically, a more principled method should be explored based on the geometry of the original nonlinear system, e.g., the nonlinear controllability indices defined through the differential-geometric/Lie-algebraic framework~\citep{isidori1985nonlinear}.

%===============================================================================

\clearpage

% The acknowledgments are automatically included only in the final and preprint versions of the paper.
\acks{
This research received funding from the European Union’s Horizon 2020 research and innovation program under the Marie Skłodowska-Curie grant agreement No. 953348 (ELO-X). 
This work was part of BrainLinks-BrainTools, which was funded by the Federal Ministry of Economics, Science and Arts of Baden-Württemberg within the sustainability program for projects of the excellence initiative II. 
This work was supported as a part of NCCR Automation, a National Centre of Competence in Research, funded by the Swiss National Science Foundation (grant number 51NF40\_225155). 
The authors thank Jasper Hoffman for the inspiring discussions and proofreading.}

%===============================================================================

\bibliography{ref}  % .bib

\newpage
\appendix

\section{Background}
\label{sec:background}

\subsection{Model Predictive Control}
\label{sec:model_predictive_control}

    Given the current state $x(t) \in \mathbb{R}^n$ at time $t$ and a planning horizon $H$, discrete-time MPC consists of a dynamic model $x_{h+1} = f_d(x_h, u_h)$, a stage-wise cost function $f_c(x_h, u_h)$, and an optimization algorithm that computes the optimal control sequences $\{u_0^*, u_1^*, \dots u_{H}^*\}$, where $u_h \in \mathbb{R}^m$. The first control input is applied, and the optimization is re-solved at the next timestep using the updated measurement. The optimization problem is formulated as follows:

 %    \begin{align}
	% \begin{split}
	%    \underset{u_0, u_1, \cdots, u_H}{\text{min}} & \quad \sum_{h=0}^{H} f_c(x_h,u_h) \\
	%    \text{s.t.} & \quad x_{h+1} =f_d(x_h, u_h) \quad x_0 = x(t).
 %    \label{eq:mpc}
	% \end{split}
 %    \end{align}
    % \vspace{-0.5em}
    \begin{equation}
        \min_{u_0, \dots, u_H} \;
        \sum_{h=0}^{H} f_c(x_h, u_h),
        \quad \text{s.t.} \;
        x_{h+1} = f_d(x_h, u_h),\;
        x_0 = x(t)
    \label{eq:mpc}
    \end{equation}
    In general, both the dynamics $f_d$ and the cost function $f_c$ are nonlinear. Sequential quadratic programming (SQP) is often employed to solve Equation~\ref{eq:mpc}, but it incurs significant computational cost due to repeated solution of QPs and gradient evaluations. The terminal cost function is omitted here, as it is often unavailable in robotic tasks~\citep{howell2022predictive}.

\subsection{Linear Predictors for Nonlinear Controlled Systems via the Koopman Operator}
\label{sec:koopman_operator}

    Lifting the state space to a higher dimension plays a key role in obtaining predictions of a nonlinear dynamical system $x^+ = f_d(x, u)$ as the output of a linear one. For uncontrolled dynamical systems, this idea can be justified with the Koopman operator theory. For controlled systems, the justification requires the extension of the state space, the new definition of observables, and their residing Hilbert space. In this paper, we follow a both rigorous and practical way of generalizing the Koopman operator $\mathcal{K}$ to controlled systems as in~\citep{korda2018linear}
    \footnote{$\bm{u} = \{u_i\}_{i=0}^{\infty}$ denotes the control sequence till infinity, $\mathcal{S}$ is the left shift operator.}: 
    
    % $\psi$ denotes the observable w.r.t the extended state $[
    %     x^T, \bm{u}^T
    % ]$ with special functional form $\psi([
    %     x^T, \bm{u}^T
    % ]) = [
    %     \phi(x)^T, \bm{u}(0)^T
    % ]$. $\mathcal{K}$ denotes the Koopman operator acting on the Hilbert space containing $\psi$. }:

    % \vspace{-0.5em}
    \begin{equation}
        \mathcal{K} \begin{bmatrix}
            \phi(x) \\
            \bm{u}(0)
        \end{bmatrix} = \begin{bmatrix}
            \phi(f(x, \bm{u}(0))) \\
            (\mathcal{S} \bm{u}) (0)
        \end{bmatrix} = \begin{bmatrix}
            \phi(f_d(x, \bm{u}(0))) \\
            \bm{u} (1)
        \end{bmatrix}, \bm{u}(0) = u.
    \label{eq:koopman-nonlinear-predictor}
    \end{equation}
    % \begin{subequations}
    % \begin{align}
        % \mathcal{K}\psi( \begin{bmatrix}
        %     x \\
        %     \bm{u}
        % \end{bmatrix} ) &= \psi( \begin{bmatrix}
        %     f(x, \bm{u}(0)) \\
        %     \mathcal{S} \bm{u}
        % \end{bmatrix}
        % ) \\
    %     \mathcal{K} \begin{bmatrix}
    %         \phi(x) \\
    %         \bm{u}(0)
    %     \end{bmatrix} &= \begin{bmatrix}
    %         \phi(f(x, \bm{u}(0))) \\
    %         (\mathcal{S} \bm{u}) (0)
    %     \end{bmatrix} = \begin{bmatrix}
    %         \phi(f(x, \bm{u}(0))) \\
    %         \bm{u} (1)
    %     \end{bmatrix}
    %     \label{eq:koopman-nonlinear-predictor}
    % \end{align},
    % \end{subequations}
    % \nth{1} line of~\ref{eq:koopman-nonlinear-predictor}\footnote{\nth{2} line of of~\ref{eq:koopman-nonlinear-predictor} is of no use because it is out of our interests to predict future values of the control sequence. } leads 
    This leads to the following (approximate) linear evolution in the lifted state space $z = \phi(x) \in \mathbb{R}^N$ while the control space remains unchanged: $z^+ = \phi(x^+) = A \phi(x) + B \bm{u}(0) = A z + B u$. In practice, a finite-dimensional lifted state $z$ is used to approximate the behaviour of the original nonlinear system~\citep{lusch2018deep, retchin2023koopman, mondal2024efficient}.

\section{Additional Algorithm Details}
\label{sec:additional_algorithm_details}

\begin{algorithm}[ht]
    \caption{LaLQR Parameter Identification}
        \begin{algorithmic}[1]
            \State \textbf{Input}: Nonlinear MPC's dynamic model $f_d(x_h, u_h)$, cost function $f_c(x_h, u_h)$ and optimization algorithm to generate optimal control $u_h = \text{MPC}(x_h)$, initial state distribution $\Delta_{\mathcal{X}}$
            \State \textbf{Output}: Embedding function $\phi, \psi$, quadratic cost parameters $Q, R$
            \State Initialize dataset $\mathcal{D}_\text{model}=\{\}$ 
            \State \textbf{for} episode number $e = 1,2,..., E$
                \State\quad Sample an initial state $x_0 ~\sim \Delta_{\mathcal{X}}$
                \State\quad \textbf{for} step $h = 0,2,..., H$ 
                    \State\quad\quad Generate optimal control $u_h$ at state $x_h$ with control rule $u_h = \text{MPC}(x_h)$
                    \State\quad\quad Rollout for next state $x_{h+1} = f_d(x_h, u_h)$ and cost $c_h = f_c(x_h, u_h)$
                    \State\quad\quad Store in the model dataset $\mathcal{D} \leftarrow \mathcal{D} \cup \{(x_h, u_h, c_h, x_{h+1})\}$
                \State\quad \textbf{end for}
        \State \textbf{end for} 
           \State \textbf{for} iteration $i = 1, 2, ...., I$
                \State\quad Sample $B$ samples $(x_h^b, u_h^b, c_h^b, x_{h+1}^b)_{b=1}^B$ from the dataset $\mathcal{D}$
                \State\quad Update parameters $p = (\phi, \psi, Q, R, F) = p - \lambda \nabla_p \sum_{b=1}^B \mathcal{L}_\text{LSP}(x_h^b, u_h^b, x_{h+1}^b) + \mathcal{L}_\text{CP}(x_h^b, u_h^b, c_h^b)$, where $\lambda$ is the learning rate
           \State \textbf{end for}
        \end{algorithmic}
    \label{alg:MPC_imitation_learning}
    \end{algorithm}    

    The pseudocode of the algorithm can be found in Algorithm~\ref{alg:MPC_imitation_learning}.
    
\section{Additional Experimental Setups}
\label{sec:additional_experimental_setups}

\subsection{Robotic Tasks on MuJoCo}
\label{sec:robotic_tasks_on_mujoco} 

    We select $4$ robotic tasks $\texttt{Cartpole}$, $\texttt{Particle}$, $\texttt{Swimmer}$ and $\texttt{Panda}$, with increasing complexity in state and action spaces as illustrated in Figure~\ref{fig:robots}. 

    \begin{itemize}
        \item \texttt{Cartpole} has an unactuated pole on top of a moving cart. In this task, the controller should balance the pole in the origin position. The state ($4$-dim) consists of the horizontal position and the angle from the vertical of the pole and their velocities. The control ($1$-dim) is the horizontal force on the cart. The cost is calculated as $c(x_t, u_t) = 0.1 \times c_\text{velocity}(x_t, u_t) + 0.1 \times c_\text{control}(x_t, u_t) + 10.0 \times c_\text{centered}(x_t, u_t) + 10.0 \times c_\text{vertical}(x_t, u_t)$. $c_\text{velocity}(x_t, u_t)$ is the absolute speed in horizontal direction. $c_\text{control}(x_t, u_t)$ is the norm of controls. $c_\text{centered}(x_t, u_t)$ is the absolute value of the position. $c_\text{vertical}(x_t, u_t)$ is the absolute value of the angle. The control frequency is $100$ Hz, and each episode runs for a horizon of $500$ steps.
        \item \texttt{Particle} is a lightweight 2-D point-mass robot controlled by applying forces along the $x$ and $y$ axes. The robot’s goal is to reach a designated target position. The state ($4$-dim) includes its 2-D position and velocity, while the control ($2$-dim) represents the planar forces applied. The cost function is defined as $c(x_t, u_t) = 5.0 \times c_\text{distance}(x_t, u_t) + 0.1 \times c_\text{velocity}(x_t, u_t) + 0.1 \times c_\text{control}(x_t, u_t)$, where $c_\text{distance}$ measures the Euclidean distance to the target, $c_\text{velocity}$ penalizes high velocities, and $c_\text{control}$ penalizes control effort. The control frequency is $100$ Hz, and each episode runs for a horizon of $500$ steps.
        \item \texttt{Swimmer} is a snake-liked robot with $5$ controllable actuators. In this task, the robot is driven to reach the target position. The state ($16$-dim) consists of the $3$-D position of the robot, the positions of $5$ actuators and their velocities. The control ($5$-dim) is the force of $5$ actuators. The cost is calculated as $c(x_t, u_t) = 10.0 \times c_\text{distance}(x_t, u_t) + 0.1 \times c_\text{control}(x_t, u_t)$. $c_\text{distance}(x_t, u_t)$ is the distance from the target position. $c_\text{control}(x_t, u_t)$ is the norm of controls. The control frequency is $100$ Hz, and each episode runs for a horizon of $500$ steps.
        \item \texttt{Panda} is a 7-DoF robotic manipulator based on the Franka Emika Panda arm. In this task, the robot must pick up a cube and place it at a fixed target location $[1.0, 1.0, 0.05]$. The state ($33$-dim) includes the cube's pose, joint positions and velocities and the gripper status. The control ($7$-dim) corresponds to the end-effector's pose (6-dim) and gripper action (1-dim). The cost function is defined as $c(x_t, u_t) = 1.0 \times c_\text{reach}(x_t, u_t) + 0.1 \times c_\text{bring}(x_t, u_t) + 0.1 \times c_\text{control}(x_t, u_t)$, where $c_\text{reach}$ measures the distance between the end-effector and the cube, $c_\text{bring}$ penalizes the deviation of the cube from the target, and $c_\text{control}$ regularizes torque magnitudes. The control frequency is $111$ Hz, and the horizon is $500$ steps.
    \end{itemize}

\subsection{Hyperparameters of Baselines}
\label{sec:hyperparameters_of_baselines}

    We list the hyperparameters for all baselines in the paper. 

\paragraph{SQP}  
    SQP requires prior knowledge of the nonlinear dynamic model $f_d$ and cost function $f_c$, which can be directly achieved from the MuJoCo platform. The optimization is calculated based on iLQG~\citep{tassa2012synthesis}, essentially a Gauss-Newton method utilizing first- and second-order derivative information. The detailed implementation follows the source code from \citet{howell2022predictive}. 

    \begin{table}[htbp]
    \centering
    \begin{tabular}{ll}
    \toprule
    \textbf{Parameter} & \textbf{Value} \\
    \midrule
    Planning horizon & $100$ \\
    Minimum line search step & $1.0 \times 10^{-3}$ \\
    Finite difference tolerance & $1.0 \times 10^{-6}$ \\
    Finite difference mode & $0$ (0: one-sided, 1: centered) \\
    Minimum regularization value & $1.0 \times 10^{-6}$ \\
    Maximum regularization value & $1.0 \times 10^{6}$ \\
    Regularization type & $0$ (0: control, 1: feedback, 2: value, 3: none) \\
    Maximum regularization iterations & $5$ \\
    \bottomrule
    \end{tabular}
    \caption{Planning and Optimization Hyperparameters of SQP}
    \label{tab:planning_hyperparams_sqp}
    \end{table}

\paragraph{CEM}  
    Similarly to SQP, CEM also requires prior knowledge of the nonlinear dynamic model $f_d$ and cost function $f_c$, which can be directly achieved from the MuJoCo platform. The detailed implementation follows the source code from \citet{howell2022predictive}. 

    \begin{table}[htbp]
        \centering
        \begin{tabular}{ll}
        \toprule
        \textbf{Parameter} & \textbf{Value} \\
        \midrule
        Planning horizon & $100$ \\
        Initial sampling variance & $0.1$ \\
        Minimum variance & $0.01$ \\
        Number of sampled trajectories & $20$ \\
        Number of elite samples & $10$ \\
        \bottomrule
        \end{tabular}
    \caption{Planning and Optimization Hyperparameters of CEM}
    \label{tab:planning_hyperparams_cem}
    \end{table}

\paragraph{LoLQR}
    LoLQR expands the nonlinear dynamic model at the stable point $(x_S, u_S)$ with Taylor expansion, and uses the ground truth cost function $f_c$ to calculate control laws. The stable point can be achieved by the final solution of the SQP controller. 
    
\paragraph{IL}
    IL directly learns a control policy $u_h = \pi(x_h)$ by imitating the output of SQP with a neural network. The network structure is set as a $3$-layer feed-forward neural network with $[512, N]$ hidden units, Mish~\citep{misra2020mish} activation function. The model is trained for $50$ epochs with $128$ batch size, and the optimizer is AdamW~\citep{loshchilov2018decoupled} with a learning rate of $0.001$. 
    The dimension $N$ of the latent space for each task can be referred to Table~\ref{tab:hidden_size}. 
    
\paragraph{LaLQR (Ours)}
    Most hyperparameters and training datasets of LaLQR are set equally as IL for a fair comparison. The embedding function $\phi$ is implemented as a 2-layer neural network, with a one-layer feed-forward neural network with $[512]$ hidden units, Mish~\citep{misra2020mish} activation function. The output dimension is $N$ so that the gain matrix $K$ has the size $N \times m$, which has similar parameters as the last layer of the policy network in IL. The model is updated for $50$ epochs with $128$ batch size, and the optimizer is AdamW~\citep{loshchilov2018decoupled} with a learning rate of $0.001$. The dimension $N$ of the latent space for each task can be referred to Table~\ref{tab:hidden_size}.
    The latent space dimension is empirically set to $20m$ for $m$ control inputs to balance representational capacity and learning difficulty, as demonstrated in Section~\ref{sec:ablation_study}. This hyperparameter warrants further investigation and tuning in future work.
     
    % -----------------------
    % method comparison
    \begin{table*}[ht]
    	\begin{center}
    		\begin{tabular}{lcccc}
            \toprule
    			\textbf{Parameters} & $\texttt{Cartpole}$ & $\texttt{Particle}$ & $\texttt{Swimmer}$ & $\texttt{Panda}$ \\
            \midrule	
                latent dimension $N$ & 20 & 40 & 100 & 140 \\
            \bottomrule
                \end{tabular}
    	\end{center}
    \caption{Dimension of latent space of IL and LaLQR for all environments.}
    \label{tab:hidden_size}
    \end{table*}
    % ------------------------ 

\section{Additional Experimental Results}
\label{sec:additional_experimental_results}

\subsection{Additional Training Results}
\label{sec:additional_training_results}

\begin{figure}[ht]
        \centering
        \includegraphics[width=0.8\linewidth]{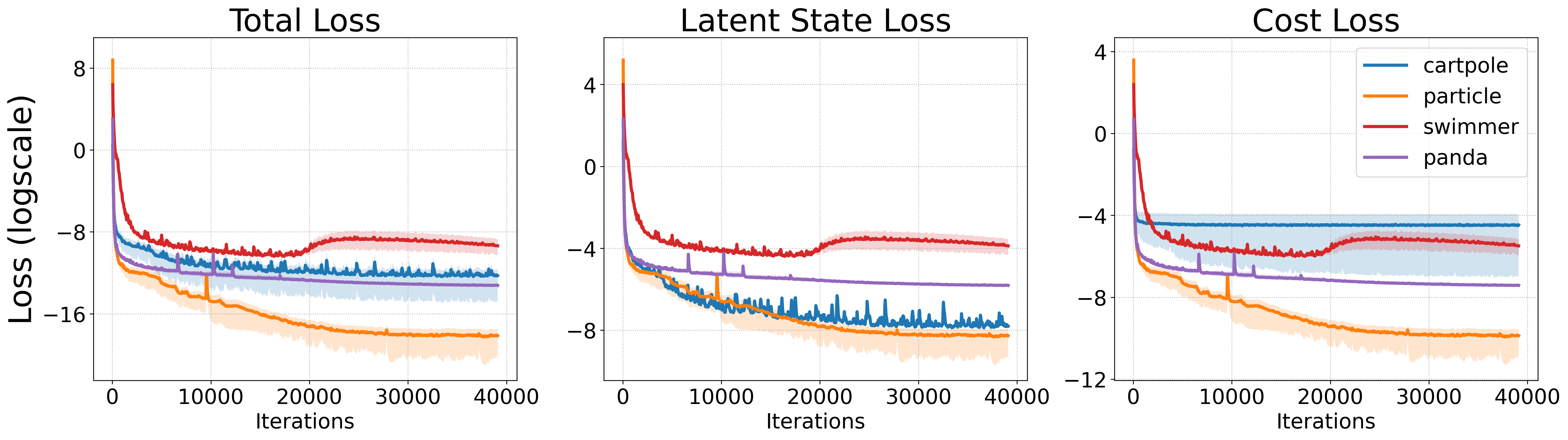}
        \caption{Training curves of LaLQR on all tasks. The a-xis is the training iterations, and the y-axis are training objectives of total loss, latent state loss and cost loss respectively. All results are plotted with the average and standard deviation shading over $3$ random seeds}
        \label{fig:train}
    \end{figure}

    We plot the training curves of LaLQR on all tasks in Figure~\ref{fig:train}. In general, this is not a difficult task for our proposed method, as all losses gradually decrease to $0$ in all tasks. 
    
\end{document}